%% file: main.tex
\documentclass[cameraready]{Interspeech}

\title{TaigiSpeech: A Low-Resource Real-World Speech Intent Dataset and Preliminary Results with Scalable Data Mining In-the-Wild}

\author[affiliation={1}, orcid=0009-0001-1562-7282]{Kai-Wei}{Chang$^{\dagger,}$}
\author[affiliation={2},orcid=0009-0007-3994-6433]{Yi-Cheng}{Lin$^{*,}$}
\author[affiliation={6},orcid=0000-0003-2125-5689]{Huang-Cheng}{Chou$^{*,}$}
\author[affiliation={2},orcid=0009-0006-6351-1088]{Wenze}{Ren$^{*,}$}
\makeatletter
\g@addto@macro\authorlist{\\}
\renewcommand{\authorsep}{}
\makeatother
\author[affiliation={2}]{Yu-Han}{Huang}
\author[affiliation={2}, orcid=0009-0001-0466-9364]{Yun-Shao}{Tsai}
\author[affiliation={2},orcid=0009-0000-0827-3271]{Chien-cheng}{Chen}
\author[affiliation={4},orcid=0000-0001-6956-0418]{Yu}{Tsao}
\author[affiliation={5},orcid=0000-0003-0191-2178]{Yuan-Fu}{Liao}
\makeatletter
\g@addto@macro\authorlist{\\}
\renewcommand{\authorsep}{}
\makeatother
\author[affiliation={6},orcid=0000-0002-1052-6204]{Shrikanth}{Narayanan}
\author[affiliation={1}]{James}{Glass}
\author[affiliation={2,3},orcid=0000-0002-9654-5747]{Hung-yi}{Lee}

\address{
    $^1$ Massachusetts Institute of Technology, USA 
    $^2$ National Taiwan University, Taipei, Taiwan \\ 
    $^3$ National Taiwan University Artificial Intelligence Center of Research Excellence, Taipei, Taiwan \\ 
    $^4$ Academia Sinica, Taiwan 
    $^5$ National Yang Ming Chiao Tung University, Taiwan \\
    $^6$ Signal Analysis and Interpretation Laboratory (SAIL), University of Southern California, USA
}

\email{kwchang@mit.edu}


\usepackage{comment}
\usepackage{booktabs}
\usepackage{tabularx}
\usepackage{array}
\usepackage{multirow} 
\usepackage{pgfplots}
\usepackage{subcaption}
\usepackage{amsmath}
\usepackage{amssymb}
\usepackage{amsfonts}
\usepackage{xcolor}
\usepackage{colortbl}
\usepackage{makecell}
\usepackage{threeparttable}
\pgfplotsset{compat=1.18}

\usepackage{CJKutf8}      
\usepackage{graphicx}    
\usepackage{longtable}    

\begin{document}

\maketitle

\begingroup
\renewcommand{\thefootnote}{\textdagger}
\footnotetext{~Project Lead. Initiated at National Taiwan University, Taiwan.}
\renewcommand{\thefootnote}{*}
\footnotetext{~Core recording team. These authors contributed equally.}
\endgroup


\begin{abstract}
Speech technologies have advanced rapidly and serve diverse populations worldwide. However, many languages remain underrepresented due to limited resources. In this paper, we introduce \textbf{TaigiSpeech}, a real-world speech intent dataset in Taiwanese Taigi (aka Taiwanese Hokkien~/~Southern Min), 
which is a low-resource and primarily spoken language.
The dataset is collected from older adults, comprising 21 speakers with a total of 3k utterances. It is designed for practical intent detection scenarios, including healthcare and home assistant applications. To address the scarcity of labeled data, we explore two data mining strategies with two levels of supervision: keyword match data mining with LLM pseudo labeling via an intermediate language and an audio-visual framework that leverages multimodal cues with minimal textual supervision. This design enables scalable dataset construction for low-resource and unwritten spoken languages. TaigiSpeech will be released under the CC BY 4.0 license to facilitate broad adoption and research on low-resource and unwritten languages. The project website and the dataset can be found on \url{https://kwchang.org/taigispeech}.
\end{abstract}

\input{Sections/Introduction}

\input{Sections/Related_works}
\input{Sections/Recorded_Dataset_v2}

\input{Sections/Audio-Visual-Mining}
\input{Sections/Experimental-Setup}
\input{Sections/Results}
\input{Sections/Conclusion}

\clearpage
\section{Acknowledgments}
We thank the Taiwanese Language and Culture Club of Keelung Community University, the Taipei City Nangang Social Welfare Center, and Chi-An Chang for their support and advice. We also acknowledge the National Center for High-Performance Computing (NIAR, Taiwan) for computing resources. This work was supported by NSTC, Taiwan (114-2917-I-564-024 to Kai-Wei Chang; 114-2917-I-564-030 to Huang-Cheng Chou), the MOE of Taiwan (Taiwan Centers of Excellence in Artificial Intelligence, through NTU AI-CoRE), US NSF (IIS 2311676), and ODNI IARPA ARTS Program (Contract D2023-2308110001).




\section{Generative AI Use Disclosure}
Generative AI was used solely for language editing; the authors take full responsibility for the manuscript's scientific content and originality.


\bibliographystyle{IEEEtran}
\bibliography{mybib}


\input{Sections/appendix}

\end{document}

%% file: Sections/Introduction.tex
\section{Introduction}
More than 7,000 languages are spoken worldwide~\cite{LanguagesoftheWorld}.
However, many of them are not supported by commercial AI services.
A substantial portion of these languages lack a universally standardized writing system, leading to inconsistent orthographic resources, or are entirely unwritten \cite{ethnologue2026}.
As a result, they remain underrepresented in modern speech and language technologies.
\emph{Southern Min (Min-nan)}, commonly referred to as \emph{Hokkien} in Southeast Asia, is one such language.
It belongs to the Min branch of Sinitic languages and is spoken by over 45 million people worldwide~\cite{ethnologue2026, kwok2018southern}.
It is used across East and Southeast Asia, as well as in diaspora communities in the Americas and beyond~\cite{LanguagesoftheWorld}.

Across regions, Southern Min exhibits substantial variation in phonology, lexicon, and accent. In Taiwan, it plays a particularly important sociolinguistic role~\cite{kubler1985influence}.
While Mandarin is the dominant language among the general population, 
\emph{Taiwanese Hokkien} (often simply referred to as \emph{Taiwanese}) remains widely spoken, especially among older adults.
As shown in Fig~\ref{fig:survey}, language usage in Taiwan exhibits clear generational differences.
Among residents aged 6 years and above (21.8M), Mandarin is the dominant primary language (66.5\%), while Taiwanese is primarily used by 31.7\% of the population.
However, this pattern reverses among older adults.
For residents aged 65 years and above (3.7M), Taiwanese becomes the dominant primary language (64.9\%), whereas only 27\% primarily use Mandarin~\cite{dgbasurvey2020}.
This demographic shift highlights the importance of supporting Taiwanese in elderly-oriented technologies and spoken language systems, especially in home assistance and emergency/healthcare scenarios.

\input{Tables/TaigiSpeech}
\input{Figures/taiwan_language_survey}
\input{Tables/slu_survey}
With rapid population ageing, an increasing proportion of older adults live alone~\cite{tseng2024examining,Lin2024SoloElderlyTaiwan,Chen_2025}.
In such circumstances, immediate access to physical devices, such as phones, may not always be feasible, particularly during sudden medical events \cite{Tran_2024,Weng_2025}.
Smart home services and voice-enabled systems, therefore, play a critical role in enabling timely access to medical assistance and community support~\cite{saripalle2024command}.
For instance, falls are among the most serious health risks faced by older adults~\cite{rubenstein2006falls,Tsai_2025}, with consequences that are especially severe for those living alone.
A critical outcome is the so-called \textit{long lie}~\cite{kubitza2023concept}, a prolonged period of immobility after a fall when an individual cannot summon help, which is associated with serious medical complications and increased mortality.
Beyond falls, other emergency conditions, such as breathing distress or acute pain, as well as daily functional needs, e.g., controlling lighting or household devices, likewise require reliable hands-free communication/systems \cite{Tsai_2017, Chithra_2025}.
These real-world scenarios underscore the need for robust spoken intent recognition systems to create a safer and more supportive living environment for elderly individuals~\cite{hamill2009development, vacher2015evaluation, vacher2019making, o2020voice}.

However, existing intent recognition datasets predominantly focus on high-resource languages~\cite{lee2024speech} such as English~\cite{lugosch2019speech}, Mandarin~\cite{zhu2019catslu}, and French~\cite{portet2019context}, leaving low-resource languages like Taiwanese Hokkien largely unexplored.
Moreover, due to limited linguistic resources, reliable automatic speech recognition (ASR) systems are often unavailable.
The lack of domain-specific datasets further constrains the development of practical spoken language understanding (SLU) systems for elderly Taiwanese speakers.
To address this gap, we introduce a real-world spoken intent dataset collected from elderly Taiwanese speakers.
We recruit participants to record natural speech across diverse healthcare and home-assistant scenarios.
The dataset comprises 8 intent categories, including 4 emergency and 4 non-emergency functional intents, forming a pivot Taiwanese Hokkien dataset for healthcare and home-assistant spoken intent recognition.

Furthermore, to address the scarcity of labeled training data, we explore two practical strategies to mine potential intent segments from in-the-wild video sources: (1) \textit{Keyword Match Mining}, which leverages subtitles in a high-resource pivot language (Mandarin), and (2) \textit{Audio-Visual Mining}, which utilizes multimodal representations to discover relevant segments without textual supervision. Together, these approaches provide scalable ways to collect training data and support the development of elderly-oriented intent recognition systems.

To assess the effectiveness of the constructed dataset, we evaluate several baseline models, including lightweight neural networks and self-supervised learning (SSL) speech models for spoken intent recognition. Experimental results show a substantial performance degradation when models trained on mined in-the-wild data are evaluated on real-world elderly recordings, indicating a significant domain mismatch. This finding highlights the challenges of real-world deployment and underscores the necessity of TaigiSpeech as a realistic benchmark for low-resource SLU.
Our contributions are summarized as follows:
\begin{itemize}
\item \textbf{TaigiSpeech Dataset.}
We introduce TaigiSpeech, a real-world spoken intent dataset collected from elderly Taiwanese Hokkien speakers in healthcare and home-assistant scenarios. The dataset comprises 8 intent categories (4 emergency and 4 functional intents), covering 21 speakers and over 3,000 utterances. It addresses a critical gap in low-resource spoken language understanding for elderly-oriented applications. The dataset will be publicly released under a CC BY 4.0 license to facilitate future research.
\item \textbf{Exploration of Data Mining Strategies.}
We investigate practical strategies for mining training data from in-the-wild sources, utilizing both cross-lingual keyword matching and audio-visual cues to address the lack of labeled data in low-resource languages.
\item \textbf{Realistic Benchmark and Empirical Insights.}
Through extensive experiments with lightweight and SSL speech models, we observe a substantial performance degradation when models trained on mined in-the-wild data are evaluated on real-world elderly recordings, indicating a significant domain mismatch. This finding highlights the challenges of real-world deployment and underscores the necessity of realistic evaluation benchmarks for low-resource spoken intent recognition.
\end{itemize}

%% file: Tables/TaigiSpeech.tex
\definecolor{lightblue}{RGB}{235,245,255}
\definecolor{lightred}{RGB}{255,235,235}

\begin{table}[t]
\centering
\vspace{-3mm}
\fontsize{8}{10}\selectfont 
\setlength{\tabcolsep}{6.5pt}
\caption{\small Statistics and intent description of TaigiSpeech dataset.}
\vspace{-3mm}
\label{tab:taigispeech_full}

\begin{tabularx}{\linewidth}{p{3cm}X}
\hline
\textbf{Category} & \textbf{Details} \\
\hline
Speakers & 21 (Male: 8, Female: 13) \\
Age Range & 54--78 (Mean: 67.9) \\
\hline

\rowcolor{lightred}
\multicolumn{2}{l}{\emph{Emergency Intents}} \\

1.~SOS\_CALL & Calls for emergency help. \\
2.~BREATH\_EMERG. & Reports breathing difficulty. \\
3.~FALL\_HELP & Indicates a fall and requests help. \\
4.~PAIN\_GENERAL & Reports physical pain or discomfort. \\

\rowcolor{lightblue}
\multicolumn{2}{l}{\emph{Non-Emergency Intents / Functional Commands}} \\

5.~CALL\_CONTACT & Requests to contact a person. \\
6.~LIGHT\_ON & Requests to turn on the lights. \\
7.~LIGHT\_OFF & Requests to turn off the lights. \\
8.~CANCEL & Cancels a previously triggered alert. \\

\hline\hline
\multicolumn{2}{l}{Total Utterances: 3,079} \\
\hline
\end{tabularx}
\end{table}

%% file: Figures/taiwan_language_survey.tex
\begin{figure}[!t]
    \centering
    \includegraphics[width=0.9\linewidth]{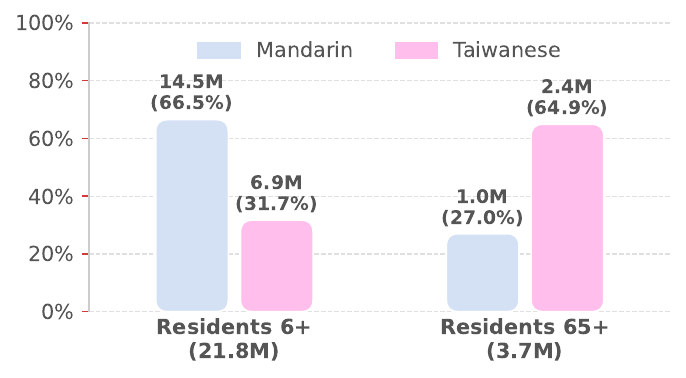}
    \vspace{-3mm}
    \caption{\small Primary language use among residents in Taiwan. Elders (aged 65+) have 64.9\% reporting Taiwanese as their primary language. Data source: 2020 Population and Housing Census, DGBAS, Taiwan~\cite{dgbasurvey2020}.}
    \label{fig:survey}
    \vspace{-4mm}
\end{figure}

%% file: Tables/slu_survey.tex
\begin{table*}[!t]
\centering
\fontsize{8}{10}\selectfont 
\setlength{\tabcolsep}{2.5pt}
\caption{\small Spoken intent and command corpora across languages and application settings. (\#~= number of; Spk. = speakers; F = female; M = male; Utt. = utterances; TTS = text-to-speech synthesized audio).}
\vspace{-3mm}
\begin{tabularx}{\textwidth}{llllllllcc}
\toprule
Corpus 
& Year
& Intent/ Command 
& \# Spk. (Gender) 
& Age 
& \#Utt. 
& Hours 
& Language 
& Smart Home 
& Emergency \\
\midrule

CEMO (Corpus 2)~\cite{devillers2005challenges}
& 2005
& Emergency
& 410 (256F; 154M)
& --
& --
& 20
& French
& --
& \checkmark \\

HIS~\cite{Fleury_2010_HIS} 
& 2010
& Emergency
& 15 (6F; 9M)
& 24-43
& 0.67k
& 0.14
& French
& \checkmark
& -- \\

CARES~\cite{young2013cares}
& 2013
& Emergency
& 40 (19M; 21F)
& 23--91
& -- 
& 53.3
& Canadian English
& --
& \checkmark \\

HIS-Distress Call~\cite{fleury2013french}
& 2013
& 20 Distress
& 10 (7M; 3F)
& $\sim$37.2
& 450
& --
& French
& --
& \checkmark \\

SWEET-HOME~\cite{Vacher_2014_SWEETHOME} 
& 2014
& Command+Emergency 
& 71 (Unknown)
& 19--91
& 9.2k
& 33
& French
& \checkmark
& \checkmark \\

ITAAL~\cite{principi2015integrated}
& 2015
& Commands+Distress
& 20 (10F; 10M)
& --
& 2.4k
& 1.1
& Italian
& \checkmark
& \checkmark \\

Speech Commands~\cite{warden2018speech} 
& 2018
& 30 keywords
& 1,881 (Unknown)
& --
& 64.7k
& 18
& English
& --
& -- \\

FSC~\cite{lugosch2019speech} 
& 2019
& 31 intents
& 97 (49F; 48M)
& --
& 30k
& 19
& English
& \checkmark
& -- \\

VocADom@A4H~\cite{portet2019context}
& 2019
& 7 intents 
& 11 (Unknown)
& $<$28
& 26k
& 12
& French 
& \checkmark
& -- \\

SLURP \cite{bastianelli2020slurp}
& 2020
& 46 actions (18 scenes)
& 177 (Unknown)
& --
& 72k
& 58
& English
& \checkmark
& -- \\

SNIPS (synthesized)~\cite{huang2020learning}
& 2020
& 7 intents
& TTS (Unknown)
& --
& 14k
& --
& English
& \checkmark
& -- \\

EMSAssist~\cite{jin2023emsassist}
& 2023
& 46 EMS protocols
& 6 (1F; 5 M)
& --
& 1.7k
& --
& English
& --
& \checkmark \\

VN-SLU~\cite{tran24b_interspeech}
& 2024
& 28 intents
& 240 (Unknown)
& --
& 17k
& 20
& Vietnamese
& \checkmark
& -- \\

MDSC~\cite{gao2024enhancing}
& 2024
& 10 wake words + others
& 46 (15F; 31M)
& 18--48
& 18.6k
& 17
& Mandarin
& \checkmark
& -- \\

\midrule
TaigiSpeech (ours)
& 2026
& 8 intents
& 21 (13F; 8M)
& 54--78
& 3,079
& 6.1
& Taiwanese Hokkien
& \checkmark
& \checkmark \\

\bottomrule
\end{tabularx}
\label{tab:intent_multilingual_corpora_full}
\vspace{-3mm}
\end{table*}

%% file: Sections/Related_works.tex
\input{Tables/taigi_corpus}

\section{Related Works}
\subsection{Spoken Language Understanding}
Spoken language understanding (SLU) aims to extract semantic meaning from speech signals and serves as a core component in voice assistants and spoken dialogue systems~\cite{arora2022espnet}. 
A central subtask in SLU is \emph{intent recognition}, which identifies the user's underlying intention from spoken utterances and triggers downstream functionality, such as information retrieval, content delivery, and device control.

Early SLU benchmarks such as ATIS~\cite{hemphill1990atis} established a narrow-domain benchmark for task-oriented query understanding, while the Speech Commands~\cite{warden2018speech} dataset later provided a widely adopted benchmark for limited-vocabulary spoken command recognition in a keyword-spotting style setting.
Subsequent datasets moved closer to practical assistant scenarios. Fluent Speech Commands (FSC)~\cite{lugosch2019speech} became a standard dataset for end-to-end spoken intent recognition in smart-home settings. 
SLURP~\cite{bastianelli2020slurp} further expanded the scale and semantic diversity of English SLU by covering multi-domain home-assistant scenarios, while SLUE~\cite{shon2022slue} promoted more systematic evaluation on natural speech and pretrained speech representations. 
In Table~\ref{tab:intent_multilingual_corpora_full}, we survey representative SLU datasets for smart home and emergency detection scenarios.

On the modeling side, conventional approaches adopt a cascade pipeline that first performs ASR and then applies text-based natural language understanding (NLU)~\cite{lugosch2019speech}. 
Although effective in high-resource settings, such pipelines are sensitive to transcription errors and depend heavily on the availability of reliable ASR systems. This limitation motivated the development of end-to-end SLU methods that predict semantic meaning and intent directly from speech signals without an intermediate text representation. 
More recently, SSL speech models~\cite{mohamed2022self} such as HuBERT~\cite{hsu2021hubert} and WavLM~\cite{chen2022wavlm} have provided informative speech representations for downstream SLU fine-tuning and can deliver strong performances~\cite{yang2024large}. 
Nevertheless, most existing datasets and methods remain centered on high-resource languages. Real-world benchmarks for elderly speakers, emergency scenarios, and low-resource unwritten languages remain scarce, which motivates the proposed TaigiSpeech dataset.





\subsection{Command/Smart-home Emergency Scenario Corpora}

The deployment of voice assistants in ambient assisted living (AAL) environments has driven the creation of specialized corpora focusing on distress calls and emergency triage. 
Unlike general smart home/command datasets \cite{warden2018speech} that primarily capture routine domestic commands (e.g., turning on lights), emergency corpora may capture the unique acoustic characteristics of high-stress situations.

An early and fundamental effort in this domain is the HIS (Health Smart Home) corpus \cite{Fleury_2010_HIS} and its targeted distress call subset \cite{fleury2013french}. 
Recorded in a fully equipped smart home laboratory in France, this dataset captures normal daily interactions alongside 20 specific distress phrases (e.g., calling for help) uttered by participants. 
While highly valuable for evaluating keyword spotting and emergency detection in noisy, reverberant home environments, the dataset primarily features younger adults (ages 22--61) and focuses on isolated distress phrases rather than complex conversational interactions. 

Moving beyond isolated keyword spotting toward conversational triage, the CARES (Canadian Adult Regular and Emergency Speech) corpus \cite{young2013cares} was developed specifically to train automatic speech recognition interfaces for personal emergency response systems (PERS). 
It features simulated emergency dialogues, isolated phrases, and spontaneous speech from 40 adult actors. 
Crucially, the CARES corpus emphasizes the voices of older adults (ages 23-91), providing essential insights into how aging and simulated physiological stress (e.g., shortness of breath, pain) affect speech recognition performance.

Despite these significant contributions, as shown in Table~\ref{tab:intent_multilingual_corpora_full}, existing emergency corpora exhibit critical limitations. 
For example, most focus on high-resource languages (French and English) and few target the elders.
To address these limitations, our proposed \textbf{TaigiSpeech} dataset bridges this gap by combining smart home automation commands with emergency distress scenarios, specifically targeting elderly speakers (ages 54--78) in a low-resource linguistic setting.

\subsection{Taiwanese Hokkien Speech Corpus}
Hokkien is a major branch of Southern Min (Min-nan) Chinese originating from Fujian, China, with over 45 million speakers worldwide, primarily in Southeast Asia (Singapore, Malaysia, Indonesia, and the Philippines), southern China, and Taiwan. 
In Taiwan, writing schemes such as Taiwanese Romanization (Tai-lo), Peh-oe-ji (POJ), and other orthographic conventions exist; however, they are not yet universally adopted, making the collection of large-scale annotated speech corpora challenging.

In recent years, efforts have been made to develop Taiwanese Hokkien speech corpora through both controlled recordings and web-crawled resources. Table~\ref{tab:taigi_corpora} summarizes representative datasets in the speech processing domain. 
As shown in the table, most existing corpora rely on read speech (e.g., TAT~\cite{liao2022taiwanese}, CommonVoice~\cite{ardila2020common}) or scripted dialogues (e.g., ML-SUPERB~\cite{shi2023ml}) designed for TTS or ASR tasks.
Some datasets are sourced from television dramas~\cite{chou2023evaluating, yang2026tg}; however, they often contain background music and post-production effects, as they are originally produced for TV broadcasting. 

While valuable for general speech recognition, these datasets lack the natural, scenario-driven expressiveness required for intent recognition in emergency and assistance contexts.
TaigiSpeech is the first Taiwanese Hokkien dataset specifically designed for spoken intent recognition, with a focus on elderly speakers. 
Collecting speech from elderly speakers is particularly valuable, as Taiwanese Hokkien is often the only fluent language spoken by many older individuals.

It is worth noting that several speech foundation models have been proposed in recent years. 
Although not specifically trained for Taiwanese Hokkien, models such as Whisper~\cite{radford2023robust} and Qwen3-ASR~\cite{shi2026qwen3} have demonstrated the ability to recognize Hokkien and other Min-nan varieties to some extent. 
We therefore include these models, cascaded with LLM, as baseline systems in our experiments.

%% file: Tables/taigi_corpus.tex
\begin{table*}[t]
\centering
\fontsize{8}{10}\selectfont 
\caption{Overview of Taiwanese Hokkien speech corpora across recording conditions and speech styles.}
\vspace{-3mm}
\begin{threeparttable}

\begin{tabular}{l l l l l l}
\toprule
\textbf{Corpus} 
& \textbf{Recording Condition} 
& \textbf{Speech Style} 
& \textbf{Target Task} 
& \textbf{Hours} 
& \textbf{Speaker Type} \\
\midrule
SuiSiann~\cite{suisiann2019}    
& Studio        
& Reading speech               
& TTS                 
& 20  
& 1 female\\

TAT~\cite{liao2022taiwanese}         
& Studio       
& Reading speech               
& ASR    
& 100
& 92 males, 108 females \\

TAT-TTS~\cite{liao2022taiwanese}
& Studio
& Reading speech
& TTS
& 40
& 2 males, 2 females \\

CommonVoice~\cite{ardila2020common}*
& Crowdsourced       
& Reading speech               
& ASR                 
& 23.67  
& Diverse speakers \\

ML-SUPERB~\cite{shi2023ml}** 
& TV production    
& Scripted dialogue    
& ASR                 
& 1.5  
& Professional Actors \\

TG-ASR~\cite{yang2026tg}      
& TV production    
& Scripted dialogue    
& ASR                 
& 30  
& Professional Actors \\

\hline 
TaigiSpeech 
& {Quiet rooms}
& Scenario-driven Expressive 
& Intent Recognition 
& 6.1  
& Elderly speakers \\
\bottomrule
\vspace{-3mm}
\end{tabular}

\begin{tablenotes}
\footnotesize
\item[*] CommonVoice refers to the Mozilla Common Voice 24.0 Taiwanese (Minnan) subset.
\item[**] Taiwanese Hokkien subset~\cite{chou2023evaluating} in Multilingual SUPERB (ML-SUPERB)
\end{tablenotes}

\end{threeparttable}
\label{tab:taigi_corpora}
\end{table*}

%% file: Sections/Recorded_Dataset_v2.tex
\section{TaigiSpeech}

In this section, we describe the data collection procedure, participant recruitment process, scenario design, and basic statistics of the TaigiSpeech dataset.

\vspace{-1mm}
\subsection{Participant Registration and Recording Interface}
\input{Figures/App}
We designed and developed a web-based recording application (Figure~\ref{fig:app}) that can be accessed on mobile phones, laptops, desktops, and tablets, using either built-in or external microphones (shown in Supplementary~\ref{apx:recording_env}). 
The application supports participant registration, speech recording, playback, and submission.

Participants first register their personal information, including (1) age (2) gender (3) education level (4) native language, (5) Taiwanese Hokkien fluency, and (6) hometown\footnote{Geographic background influences accent variation. Collecting hometown information helps ensure accent diversity and supports future phonetic and sociolinguistic studies.}.
Participants are allowed to select their preferred session duration (e.g., 30, 60, or 120 minutes, corresponding to 5, 10, or 20 utterances per intent) to reduce tiredness. 
The interface enables replaying, re-recording, reordering, and undoing recordings via mouse, touch, or keyboard input. 
We observed that providing multiple input options was essential, particularly for elderly participants who may be less familiar with certain devices, such as a mouse or touchscreen.

\subsection{Scenario Design}
TaigiSpeech targets 8 intents, covering both emergency and non-emergency functional speech commands. 
The selected intents are based on key priorities in geriatric care and common home assistance needs for aging in place.

We identified four critical emergency categories based on eldercare literature and public health statistics: \textit{Fall}, \textit{Pain}, \textit{Breathing}, and \textit{Help}. Falls are the leading cause of injury and accidental death among older adults~\cite{rubenstein2006falls}, making their detection a primary safety requirement. 
Similarly, symptoms such as acute chest pain and shortness of breath are frequent indicators of medical emergencies requiring immediate intervention~\cite{parshall2012dyspnea}. 
The generic \textit{Help} intent was included to capture diverse distress situations not covered by specific medical descriptions.
\input{Tables/TaigiSpeech-stats}
\input{Figures/TaigiSpeech-age}
\input{Figures/ambient_noise}
\input{Tables/recording_intents}
Complementing these are four non-emergency functional intents: \textit{Light on}, \textit{Light off}, \textit{Call contact}, and \textit{Cancel alert}. These were selected to assist with instrumental activities of daily living, as voice control significantly reduces the physical strain of environmental management for mobility-impaired individuals.
This balanced design (4 emergency vs. 4 functional) enables the evaluation of a system's ability to distinguish critical distress signals from routine commands, a vital feature for reliable home assistant systems.

To elicit natural speech, we design imagined scenarios that allow participants to situate themselves using Google Gemini Pro 2.5 \cite{Gemini_2025} (details, e.g. prompts, are in Supplementary~\ref{apx:content_prompt}).
For example, a prompt may describe a situation such as: ``Imagine you have fallen down the stairs and cannot get up.''
Participants are encouraged to respond freely rather than read fixed scripts to increase linguistic diversity and better reflect real-world usage.
Each intent contains 20 distinct scenarios, resulting in 160 prompts (20 x 8) per full recording session.
Examples of the imagined scenario, instruction prompt, and the participant’s produced utterance are shown in Table~\ref{tab:scenario_command}.

To further enhance realism and immersion, we provide silent video stimuli generated using Google Veo 3~\cite{Veo3_2025} (details, e.g. prompts, are in Supplementary~\ref{apx:video_prompt}).
These 10-second videos are designed to help participants better envision the scenario if needed. 
Importantly, participants are explicitly instructed that viewing these video clips was optional, and they are not required to follow the video. It serves only as a supportive context. 
Participants were free to respond in their own words, ensuring that the collected utterances reflect spontaneous and natural speech. Consequently, the resulting data includes not only direct commands but also fragmented sentences, repetitions, and urgent vocalizations typical of genuine distress.

\subsection{Recording}
The recording process was conducted in strict accordance with protocols approved by the Institutional Review Board (IRB). 
All recordings were obtained in accordance with ethical guidelines to ensure participant privacy and data protection.
During each recording session, at least one trained moderator from the research team was present to introduce the project, explain the procedures, and demonstrate how to use the recording interface.

To ensure the quality and consistency of the collected speech data, the recording interface implemented a pre-recording signal quality control protocol. 
Before each recording commenced, the system initiated a one-second silent period to measure the \emph{Ambient Noise Level}. 
The system repeatedly sampled the average frequency-domain energy, calculating a final mean value to represent the background noise. 
This indicator was logged as metadata for each audio file, providing an objective metric for assessing the recording environment's quietness and enabling data filtering for subsequent analysis.

\subsection{Statistics}
In total, we collected data from 21 elderly speakers, including 13 females and 8 males. 
Most speakers completed 160 utterances, while a small number of participants did not reach the full quota. 
On average, each speaker contributed 146.6 utterances.
The average duration per utterance is 7.15 seconds, resulting in a total of approximately 6.1 hours of recorded speech, comprising 3,079 utterances overall. All audios are recorded in mono channel with 48k Hz.
The per-intent statistics are provided in Table~\ref{tab:taigispeech_intent_stats}.
The age distribution of the participants is shown in Figure~\ref{fig:taigispeech-age} and the noise level distribution is presented in Figure~\ref{fig:noise_level}.

%% file: Figures/App.tex
\begin{figure}[!t]
    \centering
    \includegraphics[width=0.5\linewidth]{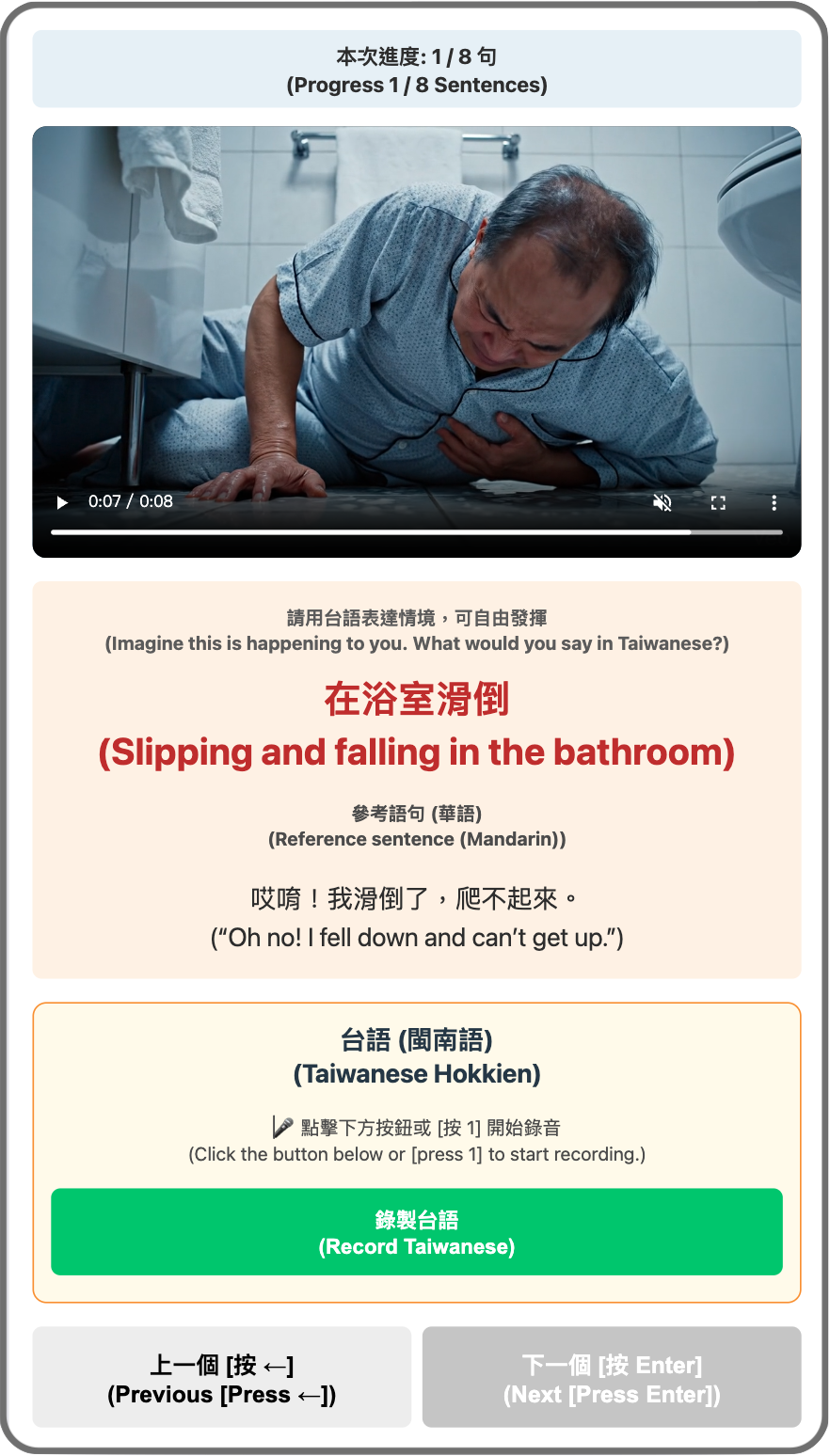}
    \caption{\small Recording app interface. The English translations are shown for illustrative purposes in this paper. All recordings were conducted under the researcher's supervision to assist elderly participants.}
    \label{fig:app}
    \vspace{-3mm}
\end{figure}

%% file: Tables/TaigiSpeech-stats.tex

\definecolor{lightblue}{RGB}{235,245,255}
\definecolor{lightred}{RGB}{255,235,235}

\begin{table}[t!]
\centering
\fontsize{8}{10}\selectfont 
\setlength{\tabcolsep}{3pt}
\caption{\small Intent-level statistics of the TaigiSpeech dataset.}
\vspace{-3mm}
\label{tab:taigispeech_intent_stats}
\renewcommand{\arraystretch}{1.15}

\begin{tabularx}{\linewidth}{lrccr}
\toprule
\textbf{Intent} & \textbf{Count} & \textbf{Duration (s)} & \textbf{Std. (s)} & \textbf{Total (min)} \\
\midrule

\rowcolor{lightred}
1.~SOS\_CALL & 387 & 7.78 & 3.28 & 50.16 \\
\rowcolor{lightred}
2.~FALL\_HELP & 385 & 7.77 & 3.73 & 49.88 \\
\rowcolor{lightred}
3.~BREATH\_EMERG & 385 & 8.22 & 3.30 & 52.71 \\
\rowcolor{lightred}
4.~PAIN\_GENERAL & 385 & 7.91 & 3.73 & 50.79 \\

\rowcolor{lightblue}
5.~CALL\_CONTACT & 385 & 7.04 & 5.30 & 45.16 \\
\rowcolor{lightblue}
6.~LIGHT\_ON & 384 & 5.95 & 2.81 & 38.08 \\
\rowcolor{lightblue}
7.~LIGHT\_OFF & 384 & 5.67 & 2.35 & 36.28 \\
\rowcolor{lightblue}
8.~CANCEL\_ALERT & 384 & 6.85 & 3.17 & 43.85 \\

\midrule
\textbf{Overall} & \textbf{3,079} & \textbf{7.15} & \textbf{3.66} & \textbf{366.91} \\
\bottomrule
\end{tabularx}
\end{table}

%% file: Figures/TaigiSpeech-age.tex
\begin{figure}[t]
    \centering
    \includegraphics[width=0.9\linewidth]{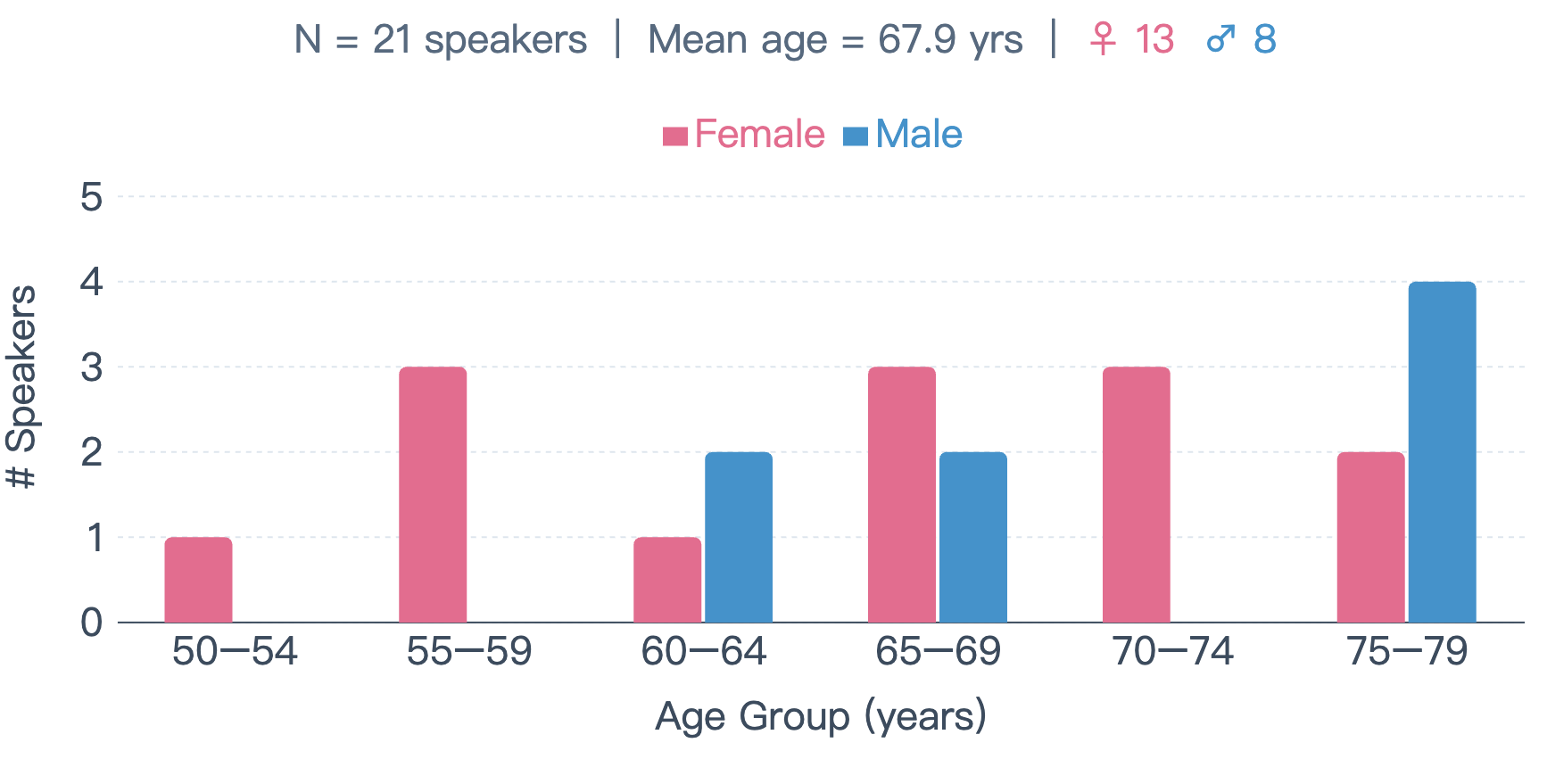}
    \vspace{-3mm}
    \caption{\small Age distribution of speakers in the TaigiSpeech corpus. Our study focuses on elderly participants.}
    \label{fig:taigispeech-age}
\end{figure}

%% file: Figures/ambient_noise.tex
\begin{figure}[t!]
    \centering
    \includegraphics[width=0.8\linewidth]{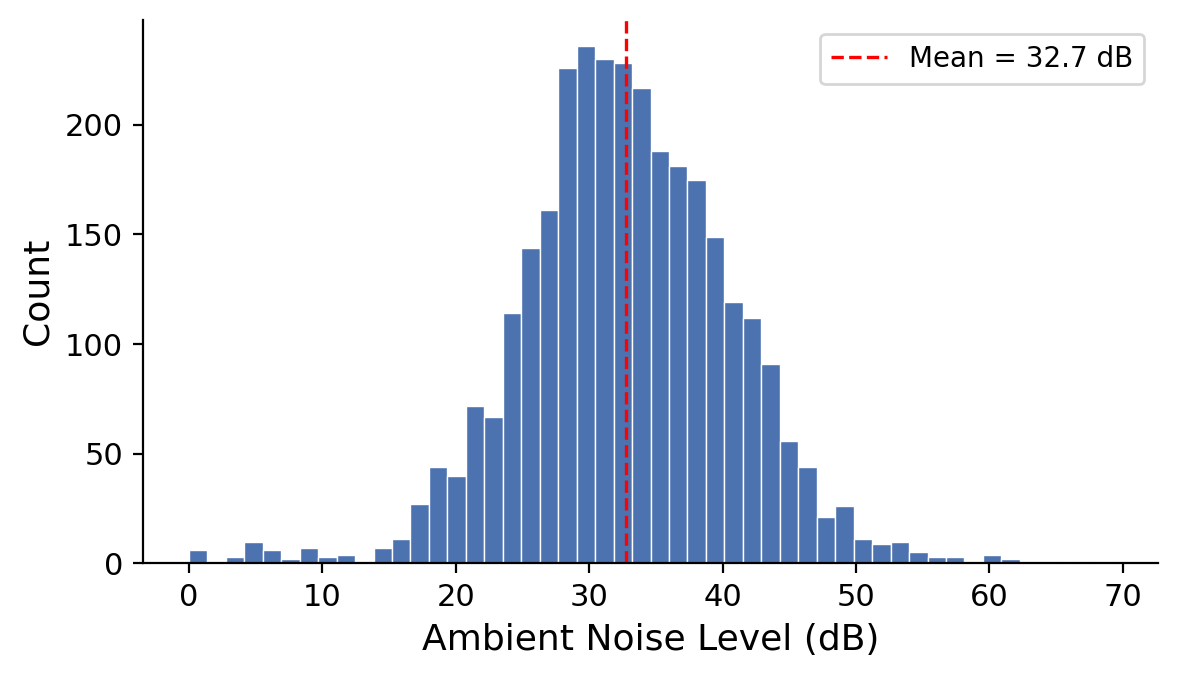}
    \vspace{-3mm}
    \caption{\small The distribution of ambient noise level (dB).}
    \label{fig:noise_level}
    \vspace{-3mm}
\end{figure}

%% file: Tables/recording_intents.tex
\begin{table*}[t]
\centering
\fontsize{8}{10}\selectfont 
\caption{\small Examples of scenario prompts and corresponding elicited spoken utterances in the TaigiSpeech recording protocol. The utterances shown here are English translations of the original Taiwanese Hokkien speech.}
\vspace{-3mm}
\label{tab:scenario_command}
\begin{tabularx}{\textwidth}{p{2.2cm}p{7.3cm}X}
\toprule
\textbf{Intent} & \textbf{Scenario Prompt (Imagined Context)} & \textbf{Example Spoken Utterance (English Translation)} \\
\midrule

\multicolumn{3}{c}{\emph{Emergency Intents}} \\
\midrule

\texttt{1.SOS\_CALL} &
You are at home and notice smoke and flames spreading. &
{``Ahh! Fire! It's on fire! Help!''} \\

\texttt{2.FALL\_HELP} &
You slipped in the bathroom and cannot stand up. &
{``Ah! I fell down... I can't get up. Please help me!''} \\

\texttt{3.BREATH\_EMERG} &
You suddenly feel chest tightness and difficulty breathing. &
{``Ah... my chest feels tight. I can't breathe well...''} \\

\texttt{4.PAIN\_GENERAL} &
You suddenly feel severe stomach pain and cannot move. &
{``Ahh... my stomach... it hurts so much. I can't move...''} \\

\midrule
\multicolumn{3}{c}{\emph{Non-emergency Functional Intents}} \\
\midrule

\texttt{5.CALL\_CONTACT} &
You feel unwell and want to call your daughter. &
{``Call my daughter... ask her to come home.''} \\

\texttt{6.LIGHT\_ON} &
You wake up at night and need more light to walk safely. &
``Turn on the bedroom light.'' \\

\texttt{7.LIGHT\_OFF} &
You are going to sleep and want to turn off the lights. &
``Turn off the lights.'' \\

\texttt{8.CANCEL\_ALERT} &
You accidentally triggered an alert and want to cancel it. &
``No, no! Cancel it. This was a mistake.''\\
\bottomrule
\end{tabularx}
\vspace{-3mm}
\end{table*}

%% file: Sections/Audio-Visual-Mining.tex
\section{Data Mining in-the-wild}
For many low-resource and unwritten languages, reliable ASR systems are often unavailable.
Also, in most cases, speech data primarily exist in web-based sources (e.g., online videos), where recordings are noisy and loosely structured.
These constraints make large-scale supervised data collection impractical, motivating the development of scalable data mining methods.
The data mining goal is to automatically discover suitable \emph{speech segments} for downstream intent recognition in the target language (e.g., Taiwanese Hokkien in this paper) from a large source pool.

Inspired by~\cite{liao2020formosa, chen2023speech}, we adopt the Taiwanese Drama dataset as a large-scale source pool for mining candidate speech segments. We select this source because its conversational style and emotional expressiveness more closely align with target smart-home and emergency scenarios than other available corpora. Furthermore, the videos are accompanied by Mandarin subtitles, which provide a practical pivot language and facilitate the design of our mining pipeline. The dataset contains about 7,000 hours of videos.
Specifically, we explore two strategies: 
(1) \textbf{Keyword Match Mining}, a stronger supervised approach that leverages Mandarin subtitles and LLM pseudo-labeling filtering; and 
(2) \textbf{Audio-Visual Mining}, a weakly supervised approach that relies on pre-trained multimodal representations with weak textual supervision.

\input{Figures/AV-pipeline}



\subsection{Keyword Match Data Mining}
Inspired by~\cite{chen2023speech}, we adopt Mandarin as an intermediate language for data mining. Although Taiwanese Hokkien does not have a widely adopted written form, Mandarin translations are usually available in the drama subtitles. 

As shown on Figure~\ref{fig:av-pipeline} (Left), for each intent $i \in \mathcal{I}$, we manually construct a set of intent-specific keywords 
$\mathcal{K}_i = \{k_1, k_2, \dots, k_m\}$. 
The initial set of keywords is constructed by native speakers proficient in both Taiwanese Hokkien and Mandarin. It is subsequently expanded to incorporate synonyms, commonly used verbal expressions, and context-dependent variations, thereby maximizing lexical coverage.
For example, for the \textsc{SOS\_CALL} intent, representative Mandarin keywords include \emph{``help''}, \emph{``call the ambulance''}, and \emph{``I need a doctor''}. 
We first retrieve subtitle sentences that contain at least one keyword from $\mathcal{K}_i$. 
For each retrieved sentence $s_t$, we further include its surrounding context (the preceding and following five sentences) to form a context window. 
An LLM is then employed to perform \emph{pseudo-labeling}, determining whether the contextualized segment semantically corresponds to the target intent $i$. 
If labeled as positive, the corresponding speech segment aligned with $s_t$ is extracted.

The statistics of the pseudo-labeled dataset are shown in Table~\ref{tab:llm_label}. 
In practice, we use at least 10 keywords per intent for keyword-based mining. 
In addition, we incorporate approximately 100 daily-life keywords (e.g., \emph{``good morning''}, \emph{``go to bed''} in Mandarin) to collect diverse non-intent samples, which are grouped into the \textsc{OTHERS} category.
It is worth noting that although the drama dataset contains approximately 7,000 hours of video, the mined data remain sparse for certain intents. 
This observation highlights the importance of constructing TaigiSpeech in order to build a robust intent recognition system for Taiwanese Hokkien in home healthcare scenarios.

Keyword Match Mining relies on subtitle translations and the LLM’s capability in the intermediate language. Although effective, it may not be applicable to unwritten languages without parallel text. To overcome this limitation, we explore an alternative audio-visual mining strategy that does not require paired textual data and operates under substantially weaker supervision.

\input{Tables/llm_label}

\input{Tables/main_results}
\subsection{Audio-Visual Data Mining}
\vspace{-1mm}
While text-aligned corpora are scarce, large amounts of audio-visual content are readily available from online videos and daily recordings. 
Acoustic signals convey paralinguistic information (e.g., emotion, urgency, prosody) that is not tied to a specific written form, and visual context provides complementary semantic grounding independent of language. 
Recent contrastive multimodal encoders~\cite{vyas2025pushing} enable cross-modal alignment in a shared embedding space without requiring explicit transcription. 
We therefore explore an audio-visual mining framework based on a pretrained cross-modal retrieval model.


Let $\mathcal{D} = \{v_j\}_{j=1}^{N}$ denote a pool of $N$ unlabeled video clips collected from available sources. 
Each clip $v_j$ consists of synchronized audio and visual streams, written as $v_j = (a_j, x_j)$, where $a_j$ is the audio signal and $x_j$ the visual frames.
Let $\mathcal{T} = \{t_i\}_{i=1}^{K}$ denote a set of intent descriptions written in a source language $L_s$, typically a high-resource language on which the multimodal encoder is pretrained. 
Our goal is to retrieve speech segments in a low-resource target language $L_t$ that correspond to each intent $t_i$, without requiring paired annotations in $L_t$.

\subsubsection{Multimodal Retrieval}
\vspace{-1mm}
We adopt a pretrained cross-modal encoder, PE-AV~\cite{vyas2025pushing}, composed of an audio-visual encoder $f_{\text{av}}(\cdot)$ and a text encoder $f_{\text{text}}(\cdot)$, mapping inputs into a shared $d$-dimensional embedding space.
For each video clip $v_j$, we compute $\mathbf{z}_j = f_{\text{av}}(v_j) \in \mathbb{R}^d$, and for each intent description $t_i$, we compute $\mathbf{q}_i = f_{\text{text}}(t_i) \in \mathbb{R}^d$.
We then measure the \emph{multimodal similarity score} using cosine similarity. 



We then retrieve the top-$M$ clips according to the similarity scores. The retrieved set is treated as pseudo-labeled data for intent $t_i$ in the target language $L_t$. This formulation enables weakly supervised cross-lingual data mining: although intents are defined in $L_s$, the shared embedding space allows retrieval of aligned speech in $L_t$ without transcription or manual labeling.

In this work we found that under the audio-visual mining setting, obtaining fine-grained emergency labels (e.g., distinguishing breath-related emergencies from fall events) remains challenging, possibly due to the cross-lingual setting and weak supervision. We instead aggregate the similarity scores of the four emergency intents and rank videos based on their summed scores to retrieve coarse-grained emergency clips. In the preliminary study, we found the pseudo-labels with this setting are consistent with that of LLM labeling.


%% file: Figures/AV-pipeline.tex
\begin{figure*}[t]
    \centering
    \includegraphics[width=0.99\linewidth]{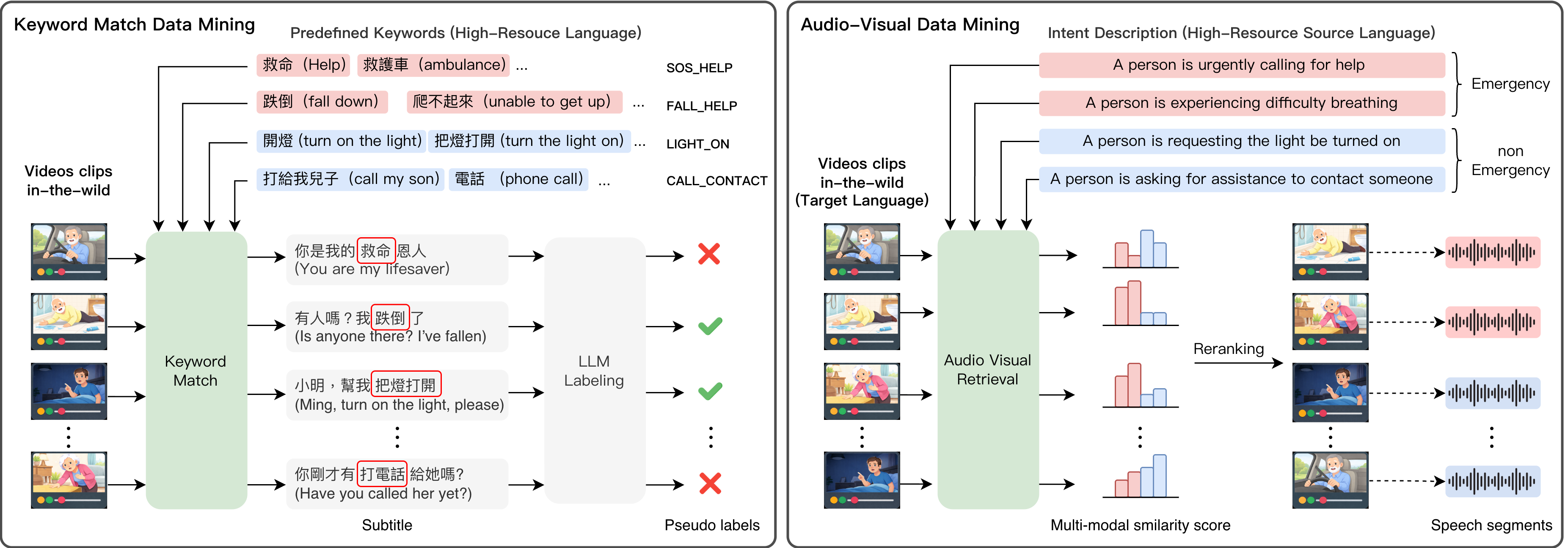}
    \vspace{-2mm}
    \caption{Overview of the explored data mining methods. Left: Keyword Match Data Mining. Right: Audio-Visual Data Mining.}
    \label{fig:av-pipeline}
    \vspace{-3mm}
\end{figure*}

%% file: Tables/llm_label.tex
\begin{table}[!t]
\centering
\fontsize{8}{10}\selectfont 
\caption{\small Statistics (number of sentences) of LLM pseudo labeling in the Keyword Match Mining method.}
\vspace{-3mm}
\label{tab:llm_label}
\begin{tabular}{lrrr}
\toprule
Intent & Total (\#) & True (\#) & False (\#) \\
\midrule
1.SOS\_CALL               & 4114 & 2801 & 1313 \\
2.FALL\_HELP              & 3017 & 165  & 2852 \\
3.BREATH\_EMERG        & 614  & 121  & 493  \\
4.PAIN\_GENERAL           & 5729 & 3054 & 2675 \\
5.CALL\_CONTACT           & 5228 & 2203 & 3025 \\
6.LIGHT\_ON               & 157  & 33   & 124  \\
7.LIGHT\_OFF              & 140  & 85   & 55   \\
8.CANCEL\_ALERT           & 946  & 55   & 891  \\
OTHERS & 7987 & 7987 & 0 \\
\bottomrule
\end{tabular}
\vspace{-4mm}
\end{table}

\begin{table}[!t]
\centering
\fontsize{8}{10}\selectfont 
\caption{\small Data statistics under different mining strategies and Taigi adaptation setting.}
\vspace{-3mm}
\label{tab:data_split_statistics}

\begin{tabular}{lcc}
\toprule
& \textbf{Keyword Mining} 
& \textbf{AV Mining} \\
\midrule
\textbf{\# Target Classes} & 5 & 2 \\
\midrule
Train & 11.3k  & 3.6k  \\
Val   & 1.4k   & 400 \\
Test (Drama) & \multicolumn{2}{c}{1.4k} \\
Test (Taigi) & \multicolumn{2}{c}{960} \\
\midrule
\multicolumn{3}{c}{\textbf{Taigi-adapt (Independent-speaker Split)}} \\
\midrule
Train& \multicolumn{2}{c}{1600 (4 male, 6 female)} \\
Val                    & \multicolumn{2}{c}{119 (1 male, 4 female)} \\
Test                   & \multicolumn{2}{c}{960 (3 male, 3 female)} \\
\bottomrule
\end{tabular}%
\vspace{-6mm}
\end{table}

%% file: Tables/main_results.tex
\begin{table*}[t!]
\centering
\caption{\small Accuracy (\%) comparison on the Drama and TaigiSpeech datasets (Taigi) under two data mining strategies. 
C=5: a 5-class setting (4 emergency intents plus a \textsc{non-emergency} category). 
C=2: binary classification (\textsc{Emergency} vs.\ \textsc{non-emergency}). 
C=8: the original 8-intent setting in TaigiSpeech.
\textbf{Taigi-adapt} means models trained on the mined data are further fine-tuned on the training set of TaigiSpeech. \textbf{Scratch}: Models trained from scratch without being pre-trained on mined data. For the Drama test set, the ground-truth labels are obtained via LLM pseudo-labeling, thus evaluating the consistency with the LLM-generated labels.}
\resizebox{\textwidth}{!}{%
\begin{threeparttable}
\vspace{-3mm}

\renewcommand{\arraystretch}{1.15}

\begin{tabular}{llccccccc}
\toprule
\multicolumn{2}{c}{\textbf{Scenario}}
& \multicolumn{3}{c}{\textbf{Keyword Match Mining}} 
& \multicolumn{3}{c}{\textbf{Audio-Visual Mining}} 
& \multicolumn{1}{c}{\textbf{Scratch}} \\
\cmidrule(lr){3-5} \cmidrule(lr){6-8} \cmidrule(lr){9-9}
Model 
& \#~Params
& Drama (C=5) 
& Taigi (C=5)
& Taigi-adapt (C=8)
& Drama (C=2) 
& Taigi (C=2)
& Taigi-adapt (C=8)
& Taigi (C=8) \\
\midrule
MatchboxNet-S
& 69k 
& 60.11{\tiny$\pm$2.51} & 45.83{\tiny$\pm$7.97} & 28.02{\tiny$\pm$5.68}
& 71.99{\tiny$\pm$2.30} & 51.88{\tiny$\pm$6.67} & 27.29{\tiny$\pm$5.73}
& 23.65{\tiny$\pm$5.99} \\
MatchboxNet-M
& 85k 
& 64.99{\tiny$\pm$2.44} & 48.02{\tiny$\pm$7.55} & 33.33{\tiny$\pm$6.20}
& 74.12{\tiny$\pm$2.37} & 52.71{\tiny$\pm$7.24} & 25.94{\tiny$\pm$5.62}
& 19.17{\tiny$\pm$4.64} \\
MatchboxNet-L
& 119k 
& 63.86{\tiny$\pm$2.51} & 43.44{\tiny$\pm$8.54} & 29.38{\tiny$\pm$6.82}
& 72.70{\tiny$\pm$2.33} & 53.75{\tiny$\pm$7.66} & 24.79{\tiny$\pm$5.68}
& 23.65{\tiny$\pm$5.47} \\
HuBERT-base        
& 94M  
& 89.60{\tiny$\pm$1.56} & 67.92{\tiny$\pm$7.03} & 90.10{\tiny$\pm$3.85}
& 79.49{\tiny$\pm$2.05} & 54.58{\tiny$\pm$8.44} & 86.77{\tiny$\pm$4.43}
& 84.69{\tiny$\pm$5.21} \\
HuBERT-large       
& 315M 
& 90.59{\tiny$\pm$1.49} & 69.48{\tiny$\pm$6.82} & 84.06{\tiny$\pm$4.79}
& 78.93{\tiny$\pm$2.05} & 53.75{\tiny$\pm$7.97} & 77.60{\tiny$\pm$5.05}
& 78.65{\tiny$\pm$5.73} \\
WavLM-base         
& 94M  
& 87.77{\tiny$\pm$1.73} & 71.25{\tiny$\pm$6.61} & 88.96{\tiny$\pm$4.64}
& 78.50{\tiny$\pm$2.12} & \textbf{54.79{\tiny$\pm$8.65}} & \textbf{89.06{\tiny$\pm$5.10}}
& \textbf{88.54{\tiny$\pm$4.58}} \\
WavLM-base-plus    
& 94M  
& 89.67{\tiny$\pm$1.59}
& \textbf{73.23{\tiny$\pm$6.04}} & \textbf{90.21{\tiny$\pm$4.12}}
& 80.55{\tiny$\pm$2.02} & 54.69{\tiny$\pm$8.54} & 87.19{\tiny$\pm$4.53} 
& 86.98{\tiny$\pm$5.00} \\
WavLM-large        
& 315M 
& \textbf{92.36{\tiny$\pm$1.38}} & 70.00{\tiny$\pm$6.93} & 83.96{\tiny$\pm$4.95}
& \textbf{81.75{\tiny$\pm$2.05}} & 52.50{\tiny$\pm$7.55} & 64.38{\tiny$\pm$6.61}
& 77.08{\tiny$\pm$5.57} \\
\bottomrule
\end{tabular}

\begin{tablenotes}
\footnotesize
\item Confidence intervals are computed using the bootstrapping method~\cite{Confidence_Intervals}.
\end{tablenotes}

\end{threeparttable}
}
\vspace{-6mm}
\label{tab:combined_results}
\end{table*}

%% file: Sections/Experimental-Setup.tex
\input{Tables/train_datasplit}

\section{Experiments}
\vspace{-1mm}
\subsection{Experimental Setup}
\vspace{-1mm}
For pseudo-labeling in the Keyword Mining approach, we employ Gemini-3 to generate intermediate annotations. For audio-visual data mining, we adopt the PE-AV large~\cite{vyas2025pushing} to extract multimodal representations and identify candidate segments without relying on paired textual supervision. Note that the PE-AV model is mainly trained on English videos and text and not trained on the low-resource Taiwanese Hokkien data. 
Due to computational constraints, rather than directly utilizing the full 7,000 hours of the drama dataset, we conduct a smaller-scale exploration using a dataset derived from keyword-matched queries to ensure that the video clips contain speech. This data pool contains \emph{unlabeled}, mixed emergency and non-emergency (other) segments, totaling about 28k video clips. Future work will focus on improving computational efficiency and conducting more comprehensive large-scale experiments. We select top 2,000 as emergency data and last 2,000 as non-emergency data.

For baseline models, we include both lightweight models (MatchboxNet~\cite{majumdar2020matchboxnet}) and self-supervised learning (SSL) speech models (HuBERT~\cite{hsu2021hubert} and WavLM~\cite{chen2022wavlm} variants), covering resource-efficient models as well as large-scale pretrained paradigms.
To construct a balanced evaluation benchmark, we select 6 speakers from TaigiSpeech to form the test set. The set consists of 3 male and 3 female speakers, with ages distributed to maintain balance across gender and age groups. Each speaker contributes 160 utterances, resulting in a total of 960 test samples.

Models are first trained on data mined using the two proposed strategies: \textbf{Keyword Match Mining} and \textbf{Audio-Visual Mining}. 
In the Keyword Match Mining setting, we focus on a 5-class classification task (four emergency intents plus a \textsc{Non-Emergency} category). 
In the Audio-Visual Mining setting, we consider a binary classification task (Emergency vs.\ Non-Emergency). 
Models trained on the mined data are evaluated on the drama dataset (denoted as \textbf{Drama}
\footnote{Pseudo labels obtained from Gemini-3 are treated as ground truth}) and the real-world recording dataset TaigiSpeech (denoted as \textbf{Taigi}).
Note that in all scenarios, only the training data differ; the testing data remain the same, and the evaluation protocol depends on the class we focus on.

To simulate a realistic low-resource domain-adaptation scenario and examine domain-mismatch effects, we further fine-tune the models pretrained in the previous stage using TaigiSpeech data of 10 speakers (totaling 1,600 utterances). The adapted models are then evaluated on the Taigi test set. This setting is denoted as \textbf{Taigi-adapt}.
The data splits used in the experiments can be found in Table~\ref{tab:data_split_statistics}.

%% file: Tables/train_datasplit.tex


%% file: Sections/Results.tex
\section{Preliminary Results}
The main results are presented in Table~\ref{tab:combined_results}. We summarize the key observations in the following sections.

\vspace{-2mm}
\subsection{Domain mismatch}
\vspace{-1mm}
A clear performance discrepancy is observed between evaluations on the drama dataset and on the real-world TaigiSpeech recordings. 
This trend consistently appears in both settings: 
\textbf{Drama (C=5)} $\rightarrow$ \textbf{Taigi (C=5)} under the Keyword Match Mining scenario, and \textbf{Drama (C=2)} $\rightarrow$ \textbf{Taigi (C=2)} under the Audio-Visual Mining scenario. 
In both cases, model performance drops substantially when transferred to the TaigiSpeech test set. 
For example, under Keyword Match Mining, the top-performing WavLM-large model drops from 92.36\% on Drama to 70.00\% on TaigiSpeech, while HuBERT-base degrades from 89.60\% to 67.92\%. 

We further observe that current audio-visual  mining approaches struggle to achieve non-trivial performance under direct domain transfer (\textbf{Taigi (C=2)}), yielding near-random binary classification accuracies across all models. Interestingly, we observed that although the retrieved samples exhibit high consistency with the LLM pseudo labels in the keyword mining stage, and can achieve approximately 70-80\% accuracy in drama dataset, this capability does not translate into Taigi.

We hypothesize that under the simplified two-class setting, the model may fail to learn robust semantic meaning, instead relying on superficial patterns that do not generalize across domains. Combined with the substantial domain mismatch between drama speech and elderly Taigi recordings, this leads to near-random performance.
Future work may involve fine-tuning the multimodal encoder, exploring alternative training objectives, and evaluating the approach under different class configurations and larger-scale datasets to better understand and improve its generalization ability.


However, under the \textbf{Taigi-adapt} setting, where a small amount of in-domain labeled data is available, performance can be substantially recovered. 
For instance, HuBERT-base and WavLM-base-plus achieve 90.10\% and 90.21\% accuracy, respectively, under the 8-class Taigi-adapt setting with Keyword Match Mining.
Overall, models pre-trained with Keyword Match Mining demonstrate stronger adaptation and perform better in Taigi-adapt compared to Audio-Visual Mining.

Although large-scale unlabeled corpora from drama sources can be effectively leveraged for data mining and pre-training, the significant degradation of real-world elderly recordings reveals the limitation of relying solely on in-the-wild mined data. 
This finding underscores the importance of TaigiSpeech as a realistic benchmark for evaluating and advancing practical spoken intent recognition systems in Taiwanese Hokkien.

\input{Figures/Confusion_matrix}

\subsection{Model Efficiency}
\vspace{-1mm}
A substantial performance gap is observed when comparing lightweight models (e.g., MatchboxNet) with SSL pre-trained models. 
This gap highlights a critical deployment challenge in smart home and healthcare scenarios, where edge devices often face strict computational constraints and privacy considerations. 
In such settings, large-scale models may not be deployable locally, and transmitting sensitive speech data to cloud servers is often undesirable or infeasible.

These findings reveal a practical trade-off between model capacity and deployability. 
Future work will therefore focus on developing more computationally efficient yet competitive architectures, improved training strategies, and more effective data mining approaches to better balance performance and real-world usability.

\subsection{Intent Classification Confusion Matrix}
\vspace{-1mm}
Figure~\ref{fig:confusion_matrix} presents the confusion matrices of four SSL-based models after continual fine-tuning. Among the four emergency intents, we observe cross-class confusion, as these intents might share similar acoustic characteristics and overlapping vocabulary in spoken Taiwanese. Among the functional intents, LIGHT\_ON and LIGHT\_OFF are the most commonly confused pair, as their spoken commands differ by only a single word. Since the test set is balanced with 160 samples per class, accuracy and macro F1 are closely aligned across all models.

\subsection{Foundation ASR+LLM cascaded methods}
\vspace{-1mm}
Recent studies have shown that large-scale foundation ASR models trained on extensive web data can transcribe or translate Southern Min (Taiwanese Hokkien) to a certain extent. Representative examples include Whisper~\cite{radford2023robust} and Qwen3-ASR~\cite{shi2026qwen3}. We therefore evaluate a cascaded approach for intent recognition based on these foundation models. Specifically, we first use the ASR model to transcribe TaigiSpeech recordings into text, and then employ a LLM, Qwen3 8B~\cite{yang2025qwen3}, to infer the corresponding intent label from the transcription.

Experimental results indicate that this cascaded ASR+LLM approach achieves reasonable performance. However, it still requires substantially larger model parameters and computational resources, and its performance remains inferior to end-to-end speech classification models such as HuBERT and WavLM. Note that we do not fine-tune the foundation ASR models and the LLM, which remains our future work.

\input{Tables/cascade}

%% file: Figures/Confusion_matrix.tex
\begin{figure}[t]
    \centering
    \includegraphics[width=0.99\linewidth]{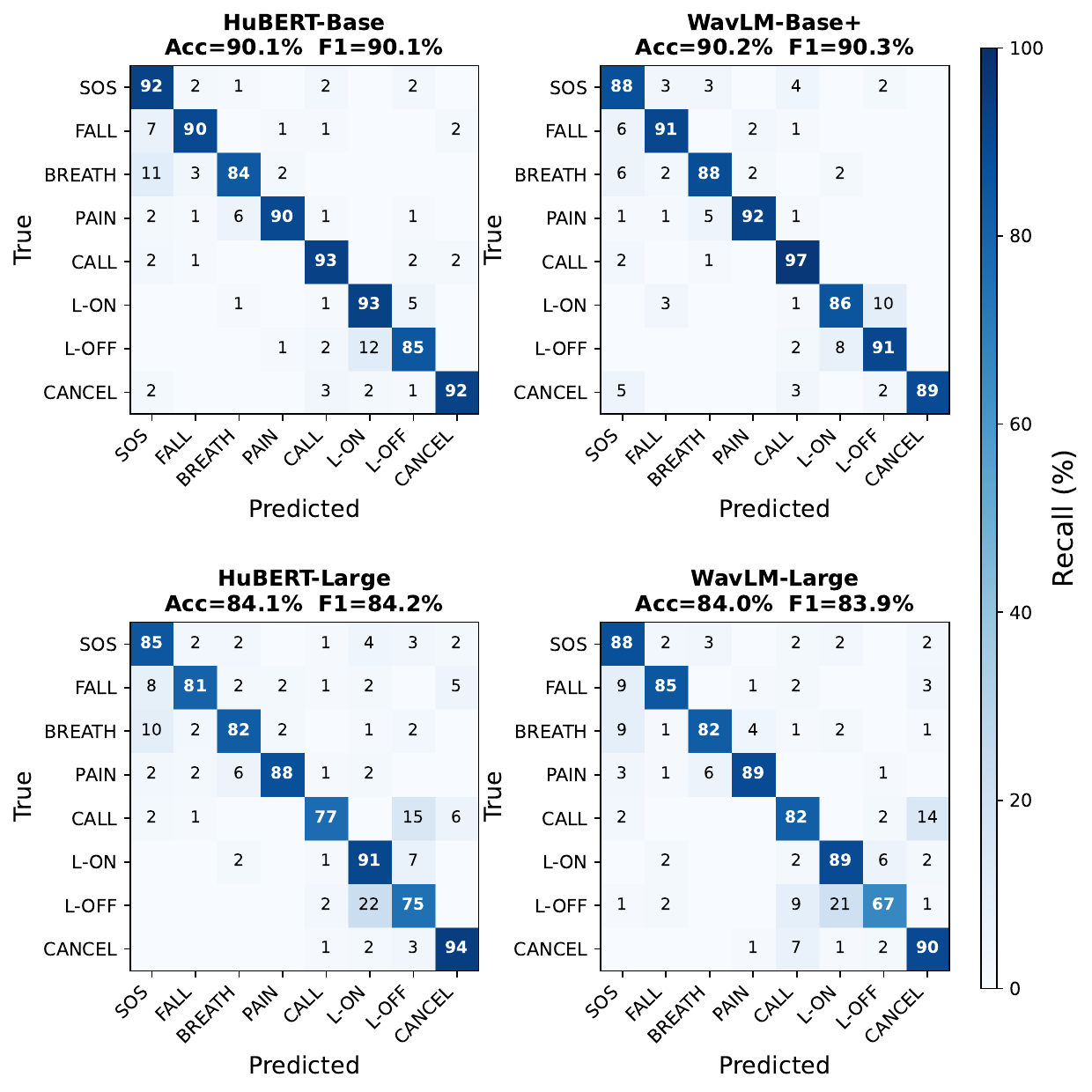}
    \vspace{-3mm}
    \caption{\small Confusion matrix of SSL speech models under the Taigi-adapt setting in the Keyword Match Mining scenario.}
    \label{fig:confusion_matrix}
    \vspace{-6mm}
\end{figure}

%% file: Tables/cascade.tex
\begin{table}[t]
\centering
\fontsize{8}{10}\selectfont 
\caption{\small Comparison between Cascade ASR+LLM and End-to-End Speech Classification Systems on Taiwanese.}
\vspace{-3mm}
\label{tab:asr_accuracy}
\begin{tabular}{lcc}
\toprule
\textbf{Model} & \textbf{\# Params} & \textbf{Accuracy (\%)} \\
\midrule
\multicolumn{3}{c}{\textit{Cascade System (ASR + Qwen3-8B)}} \\
\midrule
Whisper base      & 74M + 8B    & 15.00 \\
Whisper small      & 244M + 8B   & 15.83 \\
Whisper medium    & 769M + 8B   & 42.92 \\
Whisper large-v3   & 1.55B + 8B  & 34.38 \\
Qwen3-ASR 0.6B         & 0.6B + 8B   & 63.65 \\
Qwen3-ASR 1.7B         & 1.7B + 8B   & 74.48 \\
\midrule
\multicolumn{3}{c}{\textit{End-to-End Speech Classification}} \\
\midrule
HuBERT Base             & 94M         & 90.10 \\
WavLM Base              & 94M         & 88.96 \\
WavLM Base plus         & 94M         & \textbf{90.21} \\
\bottomrule
\end{tabular}
\vspace{-6mm}
\end{table}

%% file: Sections/Conclusion.tex
\section{Conclusion and Future Work}
\vspace{-1mm}
In this work, we introduce TaigiSpeech, a Taiwanese Hokkien spoken intent dataset designed for elderly users in home-assistant and healthcare scenarios. The dataset comprises 8 intent categories, including four emergency intents (\textsc{SOS\_CALL}, \textsc{BREATH\_EMERGENCY}, \textsc{FALL\_HELP}, and \textsc{PAIN\_GENERAL}) and four non-emergency functional intents (\textsc{CALL\_CONTACT}, \textsc{LIGHT\_ON}, \textsc{LIGHT\_OFF}, and \textsc{CANCEL\_ALERT}). 
It contains recordings from 21 elderly speakers, totaling more than 3,000 utterances. 
The dataset will be publicly released under the CC BY 4.0 license.
In addition, we explore two data mining strategies to collect training data for baseline models. 
Experimental results reveal a clear performance gap, indicating a severe domain mismatch between mined in-the-wild data and real-world elderly speech. 
This finding underscores the necessity of TaigiSpeech as a realistic benchmark for building practical Taiwanese Hokkien home-assistant systems.

In future work, we plan to investigate more efficient and robust data mining strategies and to explore lightweight yet effective speech classification models suitable for real-world deployment.
We have been expanding the dataset to include a broader range of speakers, extending beyond elderly participants in Taiwan to include Hokkien (Min-Nan) speakers worldwide. 
We hope this work will contribute to the development of inclusive spoken language technologies for low-resource languages.


%% file: Sections/appendix.tex
\clearpage
\appendix

\begin{CJK*}{UTF8}{bsmi}

\section{Supplementary Materials}

\subsection{Recording Environment}
\label{apx:recording_env}

The recording environment is illustrated by two photographs: one showing the microphone and the other depicting the recording process. 
At least one researcher was on-site to monitor and control recording quality during every recording session.

\begin{figure}[htbp]
    \centering
    \begin{subfigure}[b]{\linewidth}
        \centering
        \includegraphics[width=0.8\linewidth]{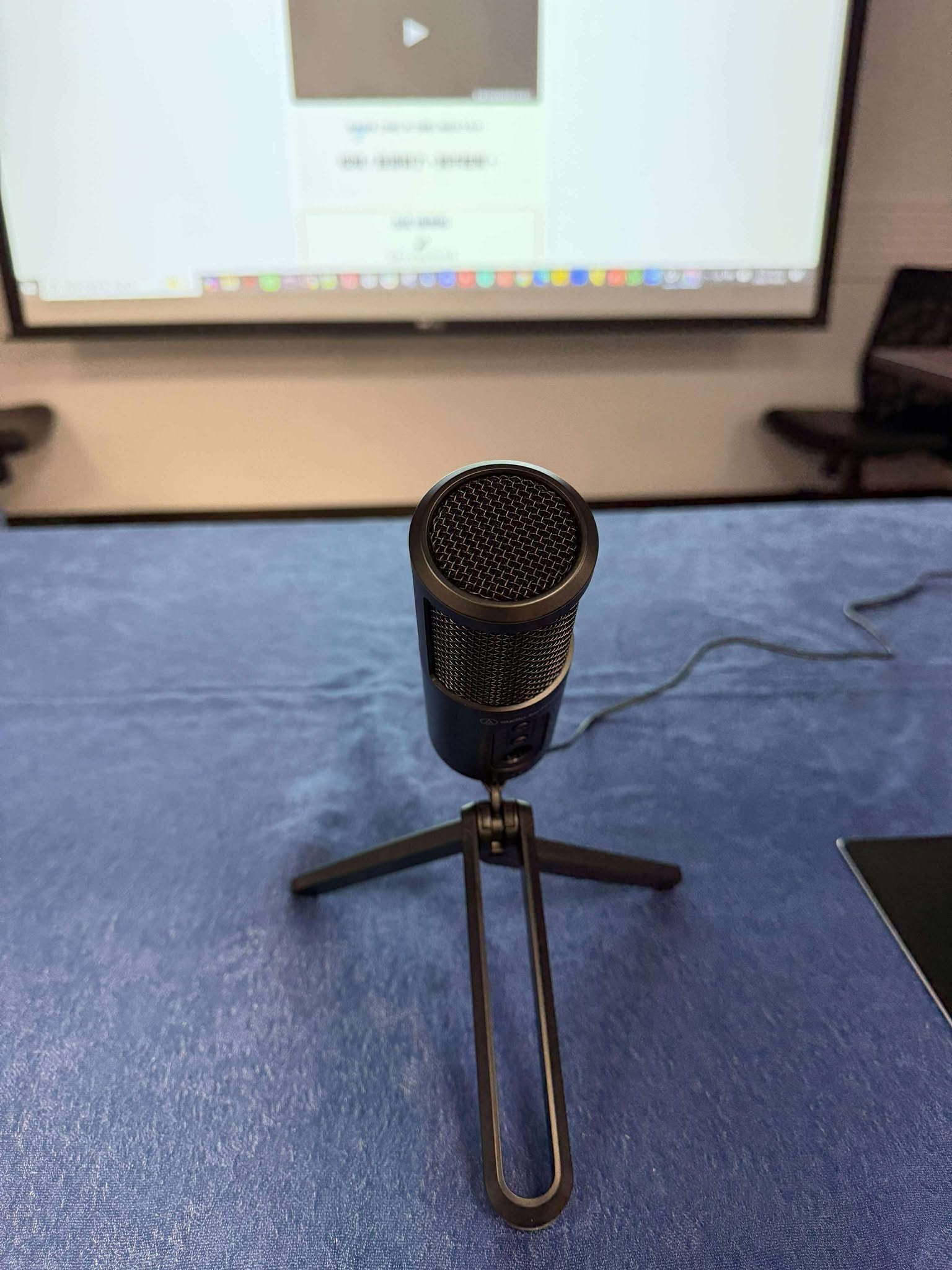}
        \caption{Microphone setup.}
        \label{fig:mic_setup}
    \end{subfigure}
    \vfill
    \begin{subfigure}[b]{\linewidth}
        \centering
        \includegraphics[width=0.8\linewidth]{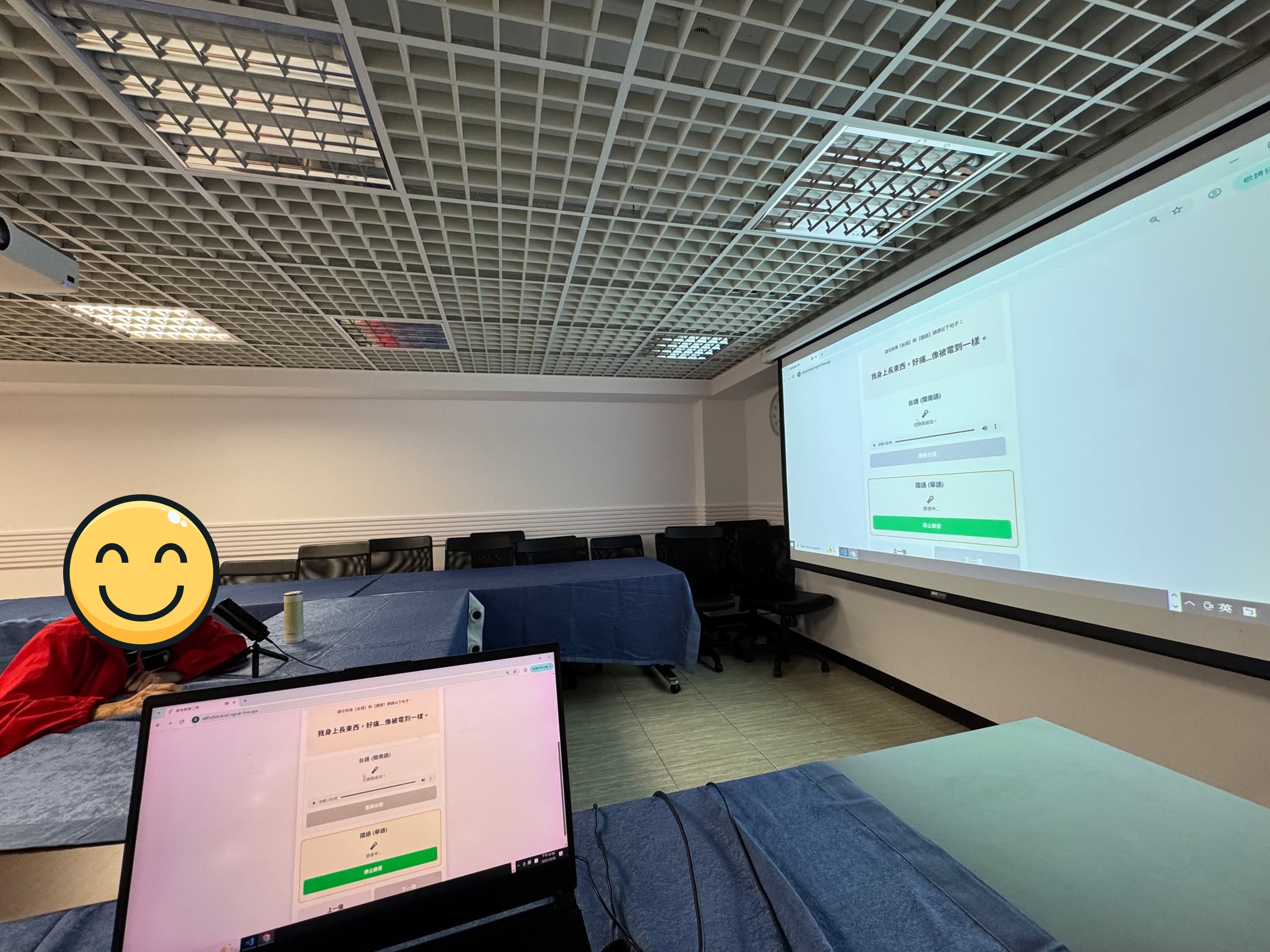}
        \caption{Recording process.}
        \label{fig:recording_process}
    \end{subfigure}
    \caption{Illustration of the recording environment.}
    \label{fig:recording_env}
\end{figure}

\subsection{Text Prompt Generation}
\label{apx:content_prompt}
We used the prompt from Google Gemini Pro 2.5 to generate the initial batch of instructions and scenarios. 
Subsequently, three researchers selected and refined the final 160 sentences. The content prompt used for this task is detailed below:

\begin{quote}
\fontsize{8}{10}\selectfont 
\textbf{Task:} Generate a JSON object including 20 entries per intent for a Taiwanese spoken language understanding scenario, including intent, description, imagined scenario, and spoken content. Your task is to create a realistic, scenario-based spoken language request from the perspective of someone in need, focusing on emergency or assistance situations. Note that situations are in-house for smart-home design. Use the provided categories and examples to guide your imagination of the situation and the corresponding vocal expression.
\newline
\textbf{Select Intent includes:}

\begin{enumerate}
\fontsize{8}{10}\selectfont 
    \item SOS\_CALL (求救)
    \item FALL\_HELP (跌倒／起不來)
    \item BREATHING\_CHEST\_EMERG (呼吸／胸悶)
    \item PAIN\_GENERAL (一般疼痛)
    \item CALL\_CONTACT (打電話)
    \item LIGHT\_ON (開燈)
    \item LIGHT\_OFF (關燈)
    \item CANCEL\_ALERT (解除警報)
\end{enumerate}

\textbf{Steps:}

\begin{enumerate}
\fontsize{8}{10}\selectfont 
    \item \textbf{Select Intent:} Choose the intent that best matches the emergency or need scenario from the provided list (e.g., SOS\_CALL, FALL\_HELP).
    \item \textbf{Describe the Situation:} Write a clear and concise description of the situation causing the need for assistance.
    \item \textbf{Imagine the Scenario:} Utilize the categories of emotions and physical states to elaborate on the imagined emotional or physical scenario.
    \item \textbf{Simulate Spoken Content:} Imagine what the person might urgently say aloud to express their need for help accurately and realistically.
    \item \textbf{Assemble JSON Output:} Combine all information into the specified JSON format.
\end{enumerate}
\end{quote}

\noindent\textbf{Output Format \& Examples:}

\begin{quote}
\ttfamily
\fontsize{8}{10}\selectfont 
Output the information as a JSON object with the following fields, \\
20 entries per intent (160 in total):\\
- intent: The selected intent indicating the type of situation.\\
- description: A brief description of the situation.\\
- imagined\_scenario: A narrative of the scenario including emotional or physical states.\\
- spoken\_content: A realistic utterance that one might say in this scenario.\\[1em]
Example:\\
\{\\
\hspace*{2ex}"index": "0001",\\
\hspace*{2ex}"intent": "SOS\_CALL",\\
\hspace*{2ex}"description": "The house is on fire (房子失火了)",\\
\hspace*{2ex}"imagined\_scenario": "Facing an immediate life threat, such as being at a fire scene, extreme panic/terrified (面臨立即的生命威脅，如火災現場，極度恐慌/魂飛魄散)",\\
\hspace*{2ex}"spoken\_content": "Help! Come save me! Help! (救命！快來救我！幫忙啊！)"\\
\}
\end{quote}

\subsection{Scenario Video Generation}
\label{apx:video_prompt}
We utilized Google Veo 3 to generate scenario videos to help subjects imagine and be reminded of the specific contexts. 
Below are the prompts used for generating some of these videos, detailed with side-by-side English and Chinese translations for clarity:

\begin{table*}[!t]
\centering
\fontsize{8}{10}\selectfont 
\renewcommand{\arraystretch}{1.3}
\caption{Scenario Video Generation Prompts (Part 1/3).}
\label{tab:video_prompts_1}
\begin{tabular}{@{} >{\raggedright\arraybackslash}p{0.18\textwidth} p{0.38\textwidth} p{0.38\textwidth} @{}}
\toprule
\textbf{Intent \& Component} & \textbf{English Prompt} & \textbf{Chinese Translation (中文翻譯)} \\
\midrule

\multicolumn{3}{@{}l}{\textbf{1. SOS\_CALL (求救)}} \\* \midrule
\textbf{Video Description}\newline(8 seconds) & 
An elderly Taiwanese man living alone in social housing in Taiwan sits on the living room sofa. Suddenly, his face turns pale, he clutches his chest tightly, and curls up his body. He opens his mouth to gasp rapidly, his eyes full of extreme panic. He struggles to reach out, trying to grab the phone on the table, but lacks strength throughout his body. Finally, his hand drops weakly, and he looks desperately towards the door. & 
一位在台灣社會住宅的獨居台灣阿公坐在客廳沙發上。突然他臉色慘白，手緊緊揪住胸口，身體蜷縮起來。他張開嘴急促地喘息，眼神中充滿了極度的慌張。他掙扎著伸出手，試圖去拿桌上的手機，但全身無力。最終虛弱地垂下手，絕望地望向門口的方向。 \\ \addlinespace

\textbf{Visual Style} & 
Realistic, with a strong sense of tension and urgency. The camera can shake slightly to simulate the viewer's nervous heartbeat, with cold color tones. & 
寫實、帶有強烈緊張與急迫感，鏡頭可以微微顫抖，模擬觀看者的緊張心跳，色調偏冷。 \\ \addlinespace

\textbf{Audio} & 
The background only has a faint television sound, quickly covered by rapid, heavy heartbeats and high-frequency tense string music. The sound fluctuates, creating a suffocating sense of life hanging by a thread. & 
背景只有微弱的電視聲，很快被一陣急促、沉重的心跳聲和高頻的緊張弦樂覆蓋。聲音時強時弱，營造出命懸一線的窒息感。 \\ \midrule

\multicolumn{3}{@{}l}{\textbf{2. FALL\_HELP (跌倒／起不來)}} \\* \midrule
\textbf{Video Description}\newline(8 seconds) & 
An elderly Taiwanese person living alone in social housing in Taiwan, wearing loungewear, tries to sit down but accidentally slips from the chair and falls heavily onto the floor. She shows a pained expression, struggles to prop herself up with both hands, and tries twice to stand up. However, her legs are visibly weak due to lack of exercise, and she eventually sits paralyzed and helpless on the spot, her eyes full of frustration and a longing for help. & 
一位住在台灣社會住宅穿著居家服的獨居台灣長者，想要坐下，不小心從椅子上滑下並重摔在地上。她臉上露出痛苦的表情，掙扎著用雙手撐地，嘗試了兩次想要站起來。但雙腿因缺乏運動明顯無力，最終只能無助地癱坐在原地，眼神充滿了挫敗與求助的渴望。 \\ \addlinespace

\textbf{Visual Style} & 
Realistic, warm home color tones. The focus is on capturing the elder's strenuous but failed attempts to stand, and the subtle facial expression changes from pain to helplessness. & 
寫實、溫馨的家居色調。重點捕捉長者嘗試站起卻失敗的吃力動作，以及臉上從痛苦轉為無助的細微表情變化。 \\ \addlinespace

\textbf{Audio} & 
Initially, quiet environmental sounds, accompanied by a heavy thud and the sound of clothing friction upon falling. After that, there is only the elder's heavy breathing from exertion and faint Taiwanese/Chinese moaning, with no background music, highlighting her isolated and helpless predicament. & 
起初是安靜的環境音，在跌倒時伴隨一聲沉重的悶響和衣物摩擦聲。之後只有長者因吃力而發出的沉重呼吸聲和細微的台灣中文呻吟，沒有背景音樂，凸顯其孤立無援的困境。 \\ \midrule

\multicolumn{3}{@{}l}{\textbf{3. BREATHING\_CHEST\_EMERG (呼吸／胸悶)}} \\* \midrule
\textbf{Video Description}\newline(8 seconds) & 
A close-up shot of the face and upper body of an elderly Taiwanese woman living alone in social housing in Taiwan. She sits quietly, then suddenly furrows her brows and presses her palm firmly against her chest. Her shoulders heave violently, her mouth opens wide, and she gasps for air laboriously like a fish out of water. Her face flushes red from hypoxia, and her eyes are filled with pain and panic. & 
近距離特寫一位在台灣社會住宅的獨居阿嬤的臉部和上半身。她靜靜坐著，突然眉頭緊鎖，手掌用力按在胸口上。她的肩膀劇烈起伏，嘴巴張得很大，像離開水的魚一樣費力地喘氣。臉上因缺氧而漲紅，眼神中滿是痛苦和恐慌。 \\ \addlinespace

\textbf{Visual Style} & 
Realistic, with a sense of oppression. Uses a shallow depth of field lens to blur the background, focusing entirely on the protagonist's pained expression and rapid movements. & 
寫實、具有壓迫感。使用淺景深鏡頭，讓背景模糊，所有焦點都集中在主角痛苦的表情和急促的動作上。 \\ \addlinespace

\textbf{Audio} & 
Absolutely no background music. Only amplified, sharp, and labored gasps, along with hissing sounds from the throat, creating an extremely uncomfortable suffocating sensation that directly impacts the viewer's hearing. & 
完全沒有背景音樂。只有被放大的、尖銳而費力的喘息聲、喉嚨間的嘶嘶聲，營造出令人極度不適的窒息感，直接衝擊觀眾的聽覺。 \\
\bottomrule
\end{tabular}
\end{table*}

\begin{table*}[!t]
\centering
\fontsize{8}{10}\selectfont 
\renewcommand{\arraystretch}{1.3}
\caption{Scenario Video Generation Prompts (Part 2/3).}
\label{tab:video_prompts_2}
\begin{tabular}{@{} >{\raggedright\arraybackslash}p{0.18\textwidth} p{0.38\textwidth} p{0.38\textwidth} @{}}
\toprule
\textbf{Intent \& Component} & \textbf{English Prompt} & \textbf{Chinese Translation (中文翻譯)} \\
\midrule

\multicolumn{3}{@{}l}{\textbf{4. PAIN\_GENERAL (一般疼痛)}} \\* \midrule
\textbf{Video Description}\newline(8 seconds) & 
An elderly Taiwanese man living alone in social housing in Taiwan is sitting in the living room watching television. He suddenly stops his actions, closes his eyes tightly, and presses one hand forcefully against his temple. His facial muscles are tense, showing that he is enduring a severe headache. He slowly puts down his book, supports his forehead with his other hand as well, and leans slightly forward. & 
一位居在台灣社會住宅的獨居阿公正坐在客廳看電視機。他突然停下動作，緊閉雙眼，一隻手用力地按壓著自己的太陽穴。他的臉部肌肉緊繃，顯現出正在忍受劇烈頭痛的樣子。他緩慢地放下書，另一隻手也扶住額頭，身體微微前傾。 \\ \addlinespace

\textbf{Visual Style} & 
Realistic, everyday life. Soft lighting, capturing the tense facial details of the protagonist caused by pain through a close-up shot. & 
寫實、生活化。光線柔和，透過特寫鏡頭捕捉主角因疼痛而緊繃的臉部細節。 \\ \addlinespace

\textbf{Audio} & 
The background consists of quiet room environmental sounds (e.g., a ticking clock). When the protagonist feels pain, a very faint but piercing high-frequency tinnitus sound effect is added to simulate the feeling of a headache, which then slowly fades. & 
背景是安靜的房間環境音（如時鐘滴答聲）。當主角感到疼痛時，加入一陣非常細微但刺耳的高頻耳鳴音效，模擬頭痛時的感受，然後慢慢減弱。 \\ \midrule

\multicolumn{3}{@{}l}{\textbf{5. CALL\_CONTACT (打電話)}} \\* \midrule
\textbf{Video Description}\newline(8 seconds) & 
An elderly Taiwanese woman living alone in social housing in Taiwan, wearing reading glasses, smiles as she flips through an old photo album. She stops at a family portrait, gently stroking the child's face in the photo with her finger, her eyes filled with longing. Then, she looks up, turns her gaze to the smart speaker in the room, moves her lips slightly, and wears an expectant expression on her face. & 
一位在台灣社會住宅獨居台灣阿嬤戴著老花眼鏡，微笑著翻閱一本舊相簿。她停在一張全家福照片上，用手指溫柔地撫摸著照片中孩子的臉龐，眼神充滿思念。接著，她抬起頭，將目光轉向房間裡的智慧音箱，嘴唇微動，臉上帶著期待的表情。 \\ \addlinespace

\textbf{Visual Style} & 
Warm, bright, and soft, with warm yellow color tones. Shot with soft lighting to create a warm, nostalgic family atmosphere. & 
溫馨、明亮且柔和，色調偏暖黃。用柔光拍攝，營造出溫暖、懷舊的親情氛圍。 \\ \addlinespace

\textbf{Audio} & 
Gentle, soothing piano or guitar background music. Accompanied by the paper friction sound of flipping through the photo album. & 
輕柔、舒緩的鋼琴或吉他背景音樂。伴隨著翻閱相簿的紙張摩擦聲。 \\ \midrule

\multicolumn{3}{@{}l}{\textbf{6. LIGHT\_ON (開燈)}} \\* \midrule
\textbf{Video Description}\newline(8 seconds) & 
In the evening, the lighting in the room is dim. An elderly Taiwanese person living alone in social housing in Taiwan squints, leans forward, and struggles immensely to read the small print on a pillbox. He tries several times but still cannot see clearly; he shakes his head in frustration, then looks up at the location of the ceiling light and mouths words as if preparing to speak. & 
傍晚，房間內光線昏暗。一位在台灣社會住宅的台灣獨居長者眯著眼睛，身體前傾，非常吃力地辨認藥盒上的小字。他試了幾次都看不清楚，困擾地搖了搖頭，然後抬頭望向天花板上電燈的位置，做出一個準備說話的口型。 \\ \addlinespace

\textbf{Visual Style} & 
Realistic, starting from dim natural light; the visual graininess can be slightly heavier to emphasize the lack of light. & 
寫實，從昏暗的自然光開始，畫面顆粒感可以稍微重一點，以強調光線不足。 \\ \addlinespace

\textbf{Audio} & 
Only very quiet environmental sounds. There can be the slight friction sound of the plastic pillbox, focusing on creating a quiet atmosphere of inconvenience caused by dimness. & 
只有非常安靜的環境音。可以有藥盒塑膠的輕微摩擦聲，重點是營造一個因昏暗而不便的安靜情境。 \\
\bottomrule
\end{tabular}
\end{table*}

\begin{table*}[!t]
\centering
\fontsize{8}{10}\selectfont 
\renewcommand{\arraystretch}{1.3}
\caption{Scenario Video Generation Prompts (Part 3/3).}
\label{tab:video_prompts_3}
\begin{tabular}{@{} >{\raggedright\arraybackslash}p{0.18\textwidth} p{0.38\textwidth} p{0.38\textwidth} @{}}
\toprule
\textbf{Intent \& Component} & \textbf{English Prompt} & \textbf{Chinese Translation (中文翻譯)} \\
\midrule

\multicolumn{3}{@{}l}{\textbf{7. LIGHT\_OFF (關燈)}} \\* \midrule
\textbf{Video Description}\newline(8 seconds) & 
At night, an elderly Taiwanese person living alone in social housing in Taiwan is already lying comfortably in bed, covered with a blanket. He lets out a big yawn, rubs his eyes, and then, with tired eyes, looks up at the ceiling light that is still on, showing obvious sleepiness in his eyes. & 
夜晚，一位在台灣社會住宅的台灣獨居長者已經舒適地躺在床上，蓋好了棉被。他打了個大大的哈欠，揉了揉眼睛，然後睜著疲憊的雙眼，看著天花板上仍然亮著的燈，眼中流露出明顯的睡意。 \\ \addlinespace

\textbf{Visual Style} & 
Realistic and tranquil. The room has warm lighting, focusing on capturing the protagonist's natural, sleepy actions like yawning and rubbing eyes. & 
寫實、寧靜。房間內是溫暖的燈光，重點捕捉主角打哈欠、揉眼睛等充滿睡意的自然動作。 \\ \addlinespace

\textbf{Audio} & 
Only faint environmental sounds or distant sounds of insects chirping. The sounds of yawning and blanket friction can be heard clearly, creating a quiet pre-sleep atmosphere. & 
只有微弱的環境音，或遠處傳來的蟲鳴聲。可以清晰地聽到哈欠聲和棉被摩擦的聲音，營造安靜的睡前氛圍。 \\ \midrule

\multicolumn{3}{@{}l}{\textbf{8. CANCEL\_ALERT (解除警報)}} \\* \midrule
\textbf{Video Description}\newline(8 seconds) & 
In the kitchen, the fire detector on the wall is flashing red and emitting a continuous "beep-beep" alarm sound. An elderly Taiwanese woman living alone in social housing in Taiwan calmly takes a towel and fans the detector, wearing a slightly helpless "not again" smile. She casually stands on her tiptoes, presses the button on the detector, and the alarm stops immediately. & 
廚房裡，牆上的防火災偵測器正閃爍著紅光，並發出持續的「嗶—嗶—」警報聲。一位在台灣社會住宅的台灣獨居阿嬤從容地拿著一條毛巾，對著偵測器搧風，臉上帶著一點「又來了」的無奈笑容。她輕鬆地踮起腳尖，按下了偵測器上的按鈕，警報隨即停止。 \\ \addlinespace

\textbf{Visual Style} & 
Realistic, everyday life. The focus is on the protagonist's calm, accustomed reaction, which sharply contrasts with the piercing alarm sound and flashing red light. & 
寫實、生活化。重點在於主角鎮定、習以為常的反應，與刺耳的警報聲和閃爍的紅光形成鮮明對比。 \\ \addlinespace

\textbf{Audio} & 
The first half features loud, piercing, and repetitive electronic alarm sound effects. At the exact moment the button is pressed, the alarm stops abruptly, and the world returns to quiet, leaving only the normal operating sound of the range hood and the sizzling sound of something frying in the pan. & 
前半段是響亮、刺耳、重複的電子警報音效。在按鈕被按下的瞬間，警報聲戛然而止，世界恢復安靜，只剩下抽油煙機的正常運轉聲和鍋裡煎東西的滋滋聲。 \\
\bottomrule
\end{tabular}
\end{table*}

\end{CJK*}